\documentclass[letterpaper, 10 pt, journal, twoside]{IEEEtran}

\usepackage{graphics} 
\usepackage{graphicx}
\usepackage{epsfig} 
\usepackage{mathptmx} 
\usepackage{times} 
\usepackage{amsmath} 
\usepackage{amssymb}  
\usepackage{rotating}
\usepackage{array}
\usepackage{multirow}
\usepackage{makecell}
\usepackage{multicol}
\usepackage{booktabs}
\usepackage{makecell}
\usepackage{colortbl}
\usepackage{hyperref}
\usepackage{algorithm}
\usepackage{algorithmic}
\usepackage{bbding}

\begin{document}
%
\title{Exploring Recurrent Long-term Temporal Fusion for Multi-view 3D Perception}
\author{Chunrui Han$^{1*}$, Jinrong Yang$^{2*}$, Jianjian Sun$^{1}$, Zheng Ge$^{1}$ \\Runpei Dong$^{3}$, Hongyu Zhou$^{1}$, Weixin Mao$^{4}$, Yuang Peng$^{5}$, Xiangyu Zhang$^{1}$
\thanks{
\noindent$^{*}$Equal Contribution\\
$^{1}$Chunrui Han, Jianjian Sun, Zheng Ge, Hongyu Zhou, Xiangyu Zhang are with the Megvii Technology.
$^{2}$Jinrong Yang is with the Huazhong University of Science and Technology.
$^{3}$Runpei Dong is with the Xi’an Jiaotong University.
$^{4}$Weixin Mao is with the Waseda University.
$^{5}$Yuang Peng is with the Tsinghua University. Corresponding author: Chunrui Han. {\tt\footnotesize (e-mail: chunrui.han@vipl.ict.ac.cn)}

}
}

\markboth{IEEE Robotics and Automation Letters. Preprint Version.}
{HAN \MakeLowercase{\textit{et al.}}: Exploring Recurrent Long-term Temporal Fusion for Multi-view 3D Perception} 

\maketitle

\begin{abstract}

Long-term temporal fusion is a crucial but often overlooked technique in camera-based Bird's-Eye-View (BEV) 3D perception. Existing methods are mostly in a parallel manner. While parallel fusion can benefit from long-term information, it suffers from increasing computational and memory overheads as the fusion window size grows. Alternatively, BEVFormer adopts a recurrent fusion pipeline so that history information can be efficiently integrated, yet it fails to benefit from longer temporal frames.
In this paper, we explore an embarrassingly simple long-term recurrent fusion strategy built upon the LSS-based methods and find it already able to enjoy the merits from both sides, i.e., rich long-term information and efficient fusion pipeline.
A temporal embedding module is further proposed to improve the model's robustness against occasionally missed frames in practical scenarios. We name this simple but effective fusing pipeline VideoBEV. Experimental results on the nuScenes benchmark show that VideoBEV obtains strong performance on various camera-based 3D perception tasks, including object detection (\textbf{55.4\%} mAP and \textbf{62.9\%} NDS), segmentation (\textbf{48.6\%} vehicle mIoU), tracking (\textbf{54.8\%} AMOTA), and motion prediction (\textbf{0.80m} minADE and \textbf{0.463} EPA).

\end{abstract}


\IEEEpeerreviewmaketitle

\section{Introduction}

\IEEEPARstart{T}{emporal} fusion technique is crucial to autonomous driving systems and it has drawn growing attention in recent years. 
Many approaches for temporal feature fusion have been developed,
and the existing research in camera-based Bird's-Eye-View (BEV) 3D perception can be divided into two categories, \emph{i.e.}, \textit{parallel} fusion and \textit{recurrent} fusion.

\begin{figure}[t]
    \begin{center}
    \includegraphics[width=\linewidth]{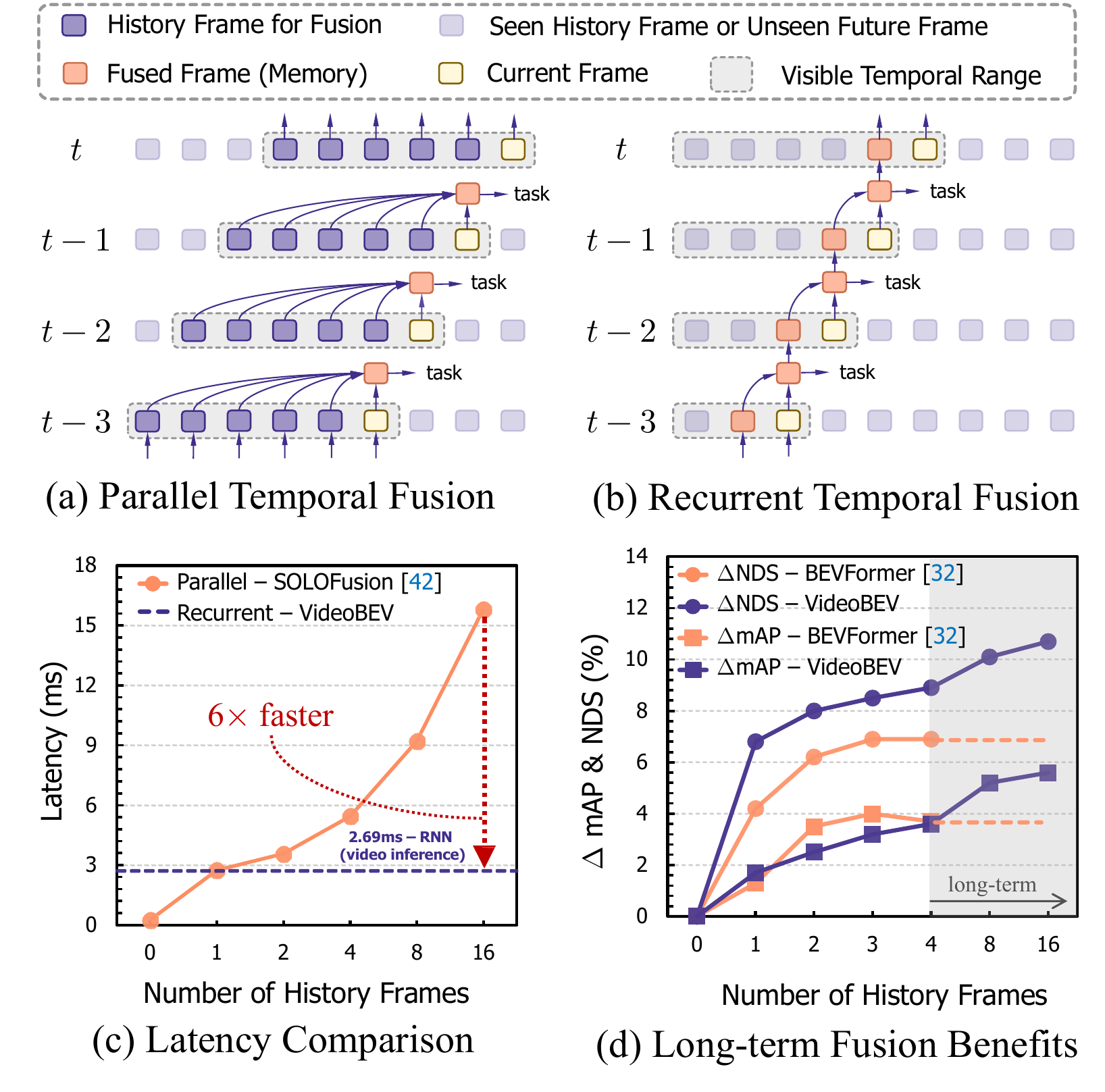}
    \vspace{-0.3in}
    \caption{\textbf{Conceptual comparison of two mainstream temporal feature fusion mechanisms}. (a) \textit{Parallel} temporal propagation within fixed temporal segments of each time stamp~\cite{VideoCNN14,TwoStreamCNNVideo14,FaF18,BEVDet4D22, SOLOFusion23,FastBEV23,BEVFormerV2}; (b) \textit{Recurrent}  temporal fusion with an iteratively updated long-term memory within the video sequence of any length~\cite{LSTM97,NeuralMachineTranslation14,Seq2SeqRNN14,BEVFormer22}. (c) Efficiency comparison between our recurrent style VideoBEV and parallel style SOLOFusion~\cite{SOLOFusion23}. (d) Comparison of benefits ($\Delta$mAP$\uparrow$ and $\Delta$NDS$\uparrow$) from long-term fusion between earlier recurrent style BEVFormer~\cite{BEVFormer22} and our VideoBEV, the numbers of BEVFormer are taken from~\cite{BEVFormer22}.} \label{fig:concept_compare}
    \end{center}
    \vspace{-0.2in}
\end{figure}

\textit{Parallel} fusion, popularized by~\cite{BEVDepth23, BEVDet4D22, PETRV2}, first aligns all history features within a fixed-length window to the current frame and then fuses them, see Fig.~\ref{fig:concept_compare}(a). This paradigm is conceptually simple but effective. A recent work~\cite{SOLOFusion23} further showcases that \textit{parallel} fusion benefits from increasing the history frame number up to 16. This covers around 8 seconds of temporal information on the nuScenes~\cite{nuScenes20} benchmark, making \textit{parallel} fusion the dominant method in this field. However, these advantages come at the cost of several issues. Firstly, \textit{parallel} fusion typically requires a \textit{fixed} window size~\cite{VideoCNN14}, which impedes the utilization of longer history frames, but real-world driving usually involves long distances. Secondly, this paradigm usually leads to a larger \textit{computation budget} compared to the \textit{recurent} manner. As shown in Fig.~\ref{fig:concept_compare}(c), SOLOFusion~\cite{SOLOFusion23} suffers from the growing latency when increasing the number of history frames. These issues hinder the application of the \textit{parallel} temporal fusion technique.

Compared to \textit{parallel} fusion, \textit{recurrent} fusion is more feasible for \textit{longer} history frames since it encodes all history information into a single memory feature (\emph{i.e.}, Fused Frame in Fig.~\ref{fig:concept_compare}(b)). However, the pioneering method BEVFormer~\cite{BEVFormer22} shows that \textit{recurrent} feature fusion cannot benefit from longer history frames. See Fig.~\ref{fig:concept_compare}(d), both mAP and NDS stop improving when the number of history frames is more than 3. The reasons could be two-fold: \textbf{(i)} the temporal fusion is \textit{intertwined} with the view transformation process of the current frame, making it more difficult to fuse temporal information, \textbf{(ii)} the spatial-temporal fusion network in BEVFormer is a Transformer~\cite{AttentionIsAllYouNeed17} architecture that is \textit{deep} and \textit{cumbersome}, which may consequently lead to the typical \textit{gradient vanishing} issue in RNN when the sequence length is long~\cite{LSTM97,GRU14}. They~\cite{BEVFormerV2} consequently turn back into the \textit{parallel} manner. As a result, in the multi-view 3D perception field, \textit{none of the existing methods can simultaneously enjoy an efficient fusing pipeline and the benefits carried by long-term information.}

Is it not feasible to apply efficient long-term temporal fusion to multi-view 3D perception tasks? The answer is \textit{no}. By leveraging a \textit{decoupled} view transformation and temporal fusion procedures on LSS-based detectors~\cite{LSS20,BEVDepth23,BEVStereo23}, we find an embarrassing fact that a simple temporal fusion strategy can facilitate our goal. During training, we sample BEV features within a sufficiently long window (\emph{e.g.}, 16 frames) and fuse them sequentially. During inference, the sequential fusion mechanism is retained throughout the entire driving process with the sampling window strategy discarded. This methodology is similar to BEVFormer~\cite{BEVFormer22}, despite that we sample more frames during training and use a framework with \textit{decoupled} spatiotemporal fusion. As a result, we obtain a simple but effective multi-frame BEV framework, dubbed VideoBEV, which can be applied to diverse perception tasks in autonomous driving.
To ensure stable and robust 3D motion perception when facing occasionally missed frames in real-world scenarios, we propose a temporal embedding module to encode timestamps, with which the dynamic temporal interval information can be effectively modeled.

Extensive experiments are conducted on four 3D perception tasks, including 3D object detection, map segmentation, object tracking, and object motion prediction. For example, on the nuScenes benchmark, VideoBEV achieves \textbf{55.4\%} mAP and \textbf{62.9\%} NDS on the 3D detection task, which improves \textbf{+2.9\%} mAP and \textbf{+1.9\%} NDS over the single-frame baseline. On the 3D object tracking benchmark that models object motion states, VideoBEV achieves \textbf{54.8\%} AMOTA, significantly outperforming the single-frame baseline by \textbf{+6.8\%}. While obtaining strong performance on various tasks, VideoBEV is still far more efficient than its long-term counterpart SOLOFusion~\cite{SOLOFusion23}. 
Although extremely simple, our VideoBEV is the first method that demonstrates the benefit of continuously increasing the number of history frames.
Besides, our VideoBEV has provided an in-depth understanding of the significance of efficient long-term temporal fusion while also establishing a new baseline for spatiotemporal multi-view 3D perception.

\section{RELATED WORKS}

\subsection{Camera-Based Single-Frame 3D Perception}
The majority of camera-based single-frame 3D prediction techniques in the beginning simply predicted 3D boxes from images. 
By creating a 3D box with the anticipated properties of a 3D object using a 2D box, Mousavian \emph{et al.}~\cite{boundingbox2017} pioneered this direction.
FCOS3D~\cite{FCOS3D21} simply extends the 2D object detector~\cite{FCOS19} to a 3D object detector by decoupling the defined 7-DoF 3D targets as 2D and 3D attributes. 
PETR~\cite{PETR22} encodes the position information of 3D coordinates into image features, producing the 3D position-aware feature. Inspired by LiDAR-based methods~\cite{MV3D17,PointPillars19}, recent advances employ view transformation to transform the feature from perspective view to the Bird's-Eye-View (BEV) for unified 3D detection. LSS~\cite{LSS20} proposes the LSS-based view transformation method, which first ``lift''s each image individually into a frustum of feature
for each camera, then ``splat''s all frustums into a rasterized BEV grid. BEVDet~\cite{BEVDet21} utilizes the LSS-based view transformation to extract BEV features and conducts 3D detection thereon. To achieve more trustworthy depth for LSS-based view transformation, BEVDepth~\cite{BEVDepth23} uses the depth from LiDAR as the supervision for precise depth estimation.

\subsection{Camera-Based Multi-Frame 3D Perception}
Multi-frame fusion for LiDAR-based 3D detectors is a widely used technology~\cite{zong2023temporal,Rv-fusenet,CenterPoint21}. However, 3D perception from a single vision frame without LiDAR is an ill-posed problem due to the lack of accurate depth information. Recent works make efforts to multi-frame 3D perception since different frames generally offer different views of objects. Saha \emph{et al.}~\cite{Image2BEV22} formulate BEV map construction from an image as a set of 1D sequence-to-sequence translations and propose a dynamic module incorporating temporal information from past estimation to build a spatiotemporal BEV representation. BEVDet4D~\cite{BEVDet4D22} extends BEVDet~\cite{BEVDet21} and fuses the history frame's features with the current frame after removing ego-motion impact. PETRv2~\cite{PETRV2} directly achieves the temporal alignment by simply aligning the 3D coordinates of the history and current frames. BEVFormer~\cite{BEVFormer22} designs a temporal self-attention to recurrently fuse the history BEV information for obtaining precise BEV features. BEVStereo~\cite{BEVStereo23} employs the temporal multi-view stereo (MVS)~\cite{StereoMaching94} to tackle the ill-posed issue of depth perception in camera-based 3D tasks. STS~\cite{STS22} leverages the geometry correspondence between frames across time to facilitate accurate depth learning. The above methods employ only limited history frames for temporal fusion. Differently, SOLOFusion~\cite{SOLOFusion23} aligns the BEV feature from the previous timesteps of a long history to the current timestep and concatenates them for long-term temporal fusion. However, it suffers from high inference latency, memory, and module parameter bottleneck. Our proposed recurrent temporal fusion module can avoid these issues. Besides, for the first time, our VideoBEV successfully demonstrates the benefit of continuously increasing the number of history frames.

\section{METHODOLOGY}
\label{metho}

\begin{figure*}[t!]
    \begin{center}
    \includegraphics[width=0.9\linewidth]{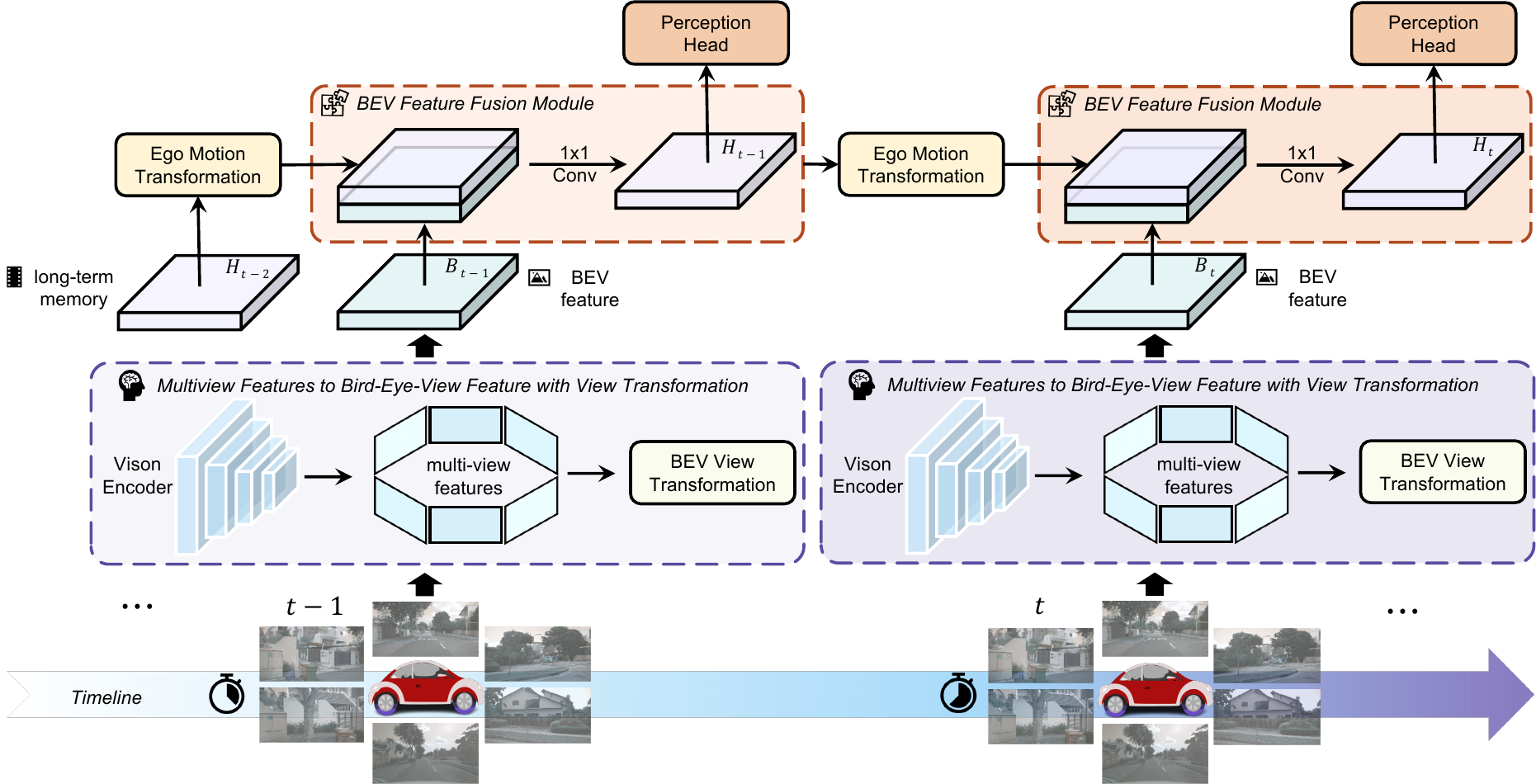}
    \vspace{-0.1in}
    \caption{\textbf{Overview of VideoBEV}. The backbone first extracts image features of different views of a frame, which are transformed to BEV from the image view to obtain the BEV feature. Then, the recurrent fusion module fuses the new BEV feature with the one of long-term memory, based on which the memory is updated and the 3D perception tasks are conducted.} \label{fig:framework}
    \end{center}
    \vspace{-0.2in}
\end{figure*}

VideoBEV employs a recurrent long-term fusion module that fuses a video stream sequentially.
A temporal embedding module is further introduced to tackle the instability of perception caused by missed frames in real-world circumstances. The overall architecture is shown in Fig.~\ref{fig:framework}. 
Sec.~\ref{sec:overview_of_videoBEV} first gives a brief overview of VideoBEV, then Sec.~\ref{sec:RNN_style_bev_fusion} introduces the recurrent style BEV fusion in detail. In the end, Sec.~\ref{sec:time_embedding} introduces the temporal embedding modeling.

\subsection{Overview of VideoBEV}
\label{sec:overview_of_videoBEV}

The overall pipeline of VideoBEV is similar to that of existing LSS-based~\cite{LSS20} 3D BEV detection, \emph{e.g.}, BEVDet~\cite{BEVDet21}, BEVDepth~\cite{BEVDepth23} and BEVStereo~\cite{BEVStereo23}, \emph{etc}, except the recurrent style temporal fusion and temporal embedding. Generally, it can be separated into three modules:
\begin{enumerate}
    \item BEV feature extraction module: a backbone network extracts the per-frame image feature of different camera views, which is further translated from perspective view to BEV for obtaining the BEV feature.
    \item Temporal fusion module: the recurrent style temporal fusion module fuses the BEV feature of the input frame with the stored long-term memory. Besides, a recurrent style temporal embedding module is employed to embed the sequence of time intervals between adjacent frames in the video sequence.
    \item 3D perception module: a 3D perception head is applied to the fused BEV feature and temporal embedding to conduct 3D perception for the input frame. 
\end{enumerate}

In the BEV feature extraction module, the backbone can be any network, \emph{e.g.}, ResNet-50~\cite{ResNet16}, ConvNeXt-B~\cite{ConvNeXt22}; the view transformation (VT) can be generally LSS-based VT such as BEVDet~\cite{BEVDet21}, BEVDepth~\cite{BEVDepth23}, and MatrixVT~\cite{MatrixVT22}, \emph{etc}, or query-based VT such as BEVFormer \cite{BEVFormer22}. We utilize the LSS-based VT due to its effectiveness. In the 3D perception module, the head can be any BEV-based task, \emph{e.g.}, 3D objection detection, map segmentation, and tracking, \emph{etc}. The temporal fusion module is newly proposed in this paper and will be introduced in the following subsections.

\subsection{Recurrent Temporal BEV Feature Fusion}
\label{sec:RNN_style_bev_fusion}

The BEV feature of a single frame generally describes objects from a single view (time step), which is inadequate for precise 3D perception. To obtain abundant features of objects, recent works, \emph{e.g.}, SOLOFusion~\cite{SOLOFusion23}, explore the temporal context information as the substitute for multi-views since different frames often offer different views of subjects.
However, as pointed out by BEVFormer~\cite{BEVFormer22} and BEVFormer V2~\cite{BEVFormerV2}, existing \textit{recurrent} style fusion fails to bring further performance gains with long-term sequence.
In contrast, the \textit{parallel} temporal fusion is able to fuse long-term video sequences effectively. Hence, we motivate our long-term recurrent style temporal fusion model from that of the sliding-window methods, introduced next.

To better understand the recurrent style fusion, we first revisit the parallel fusion with a temporal window size $k$ in SOLOFusion~\cite{SOLOFusion23}. Suppose the BEV feature of a video sequence as $\{B_{i}\}_{i=1}^{T}$, and $B_{i}$ is the BEV feature of the frame at time step $t_{i}$. The parallel fusion for the $i$-th frame in SOLOFusion \cite{SOLOFusion23} can be written as:
{\small
\begin{align}
    \hat{H}_{i} = \Big[f_{\text{sample}}(B_{i\!-\!k\!+\!1}, P_{i,i\!-\!k\!+\!1}); \dots; f_{\text{sample}}(B_{i\!-\!1},P_{i,i\!-\!1}); B_{i}\Big] *\mathbf{U},
\end{align} 
}
\noindent where $\hat{H}_{i}$ is the fused BEV feature for $i$-th frame, $P_{j, i}$ is the view transformation matrix from the ego coordinate of $j$-th frame to that of $i$-th frame considering the ego-motion, $f_{\text{sample}}$ refers to the grid sampling operation proposed by Jaderberg \emph{et al.}~\cite{STNet2015}, $\mathbf{U}$ is the convolution kernel, $[x;y]$ represents the concatenation of $x$ and $y$ along the channel dimension, and $*$ denotes the convolution operator. As can be seen, in the parallel temporal fusion, a concatenation operator is applied first to concatenate the aligned BEV feature in the window, on which the convolution operator is employed to fuse BEV features of different frames. The above formulation can be further expanded by splitting the kernel $\mathbf{U}$ along the channel dimension, \emph{i.e.},
\begin{align}
    \hat{H}_{i} &\!=\! \Big[f_{\text{sample}}(B_{i\!-\!k\!+\!1},P_{i,i\!-\!k+1});\dots;f_{\text{sample}}(B_{i\!-\!1}, P_{i\!-\!1,i}); B_{i}\Big] * \notag \\
    & \quad \Big[\mathbf{U}_{1};\dots;\mathbf{U}_{k-1};\mathbf{U}_{k}\Big] \notag \\
                &= \sum_{j=1}^{k} f_{\text{sample}}(B_{i-k+j},P_{i,i-k+j}) *\mathbf{U}_{j}, \label{sliding_window_fusion}
\end{align}
where $\mathbf{U}_{j}$ is the $j-$th chunk by equally splitting $\mathbf U$ along the channel dimension. 

The formulation of recurrent fusion is similar to that of parallel fusion. The difference is that instead of storing all the history BEV features in the temporal window and concatenating them, we store only the long-term memory of BEV feature and concatenate it with that of the current frame. 
Taking $\overline{H}_{i}$ as long-term memory of BEV feature at time step $t_{i}$, the formulation of recurrent style fusion is:
\begin{align}
    \overline{H}_{i} = \Big[f_{\text{sample}}(\overline{H}_{i-1},P_{i,i-1}); B_{i}\Big] * \mathbf{V},
\end{align}
where $\mathbf{V}$ is the convolution kernels. Considering the long-term BEV feature memory $\overline{H}_{i-1}$ is obtained by fusing the $\overline{H}_{i-2}$ to the BEV feature of $(i-1)$-th frame, the above formulation can be further expanded by splitting the kernel $\mathbf{V}$ into two chunks $\mathbf{V}_{\text{mem}}$ and $\mathbf{V}_{\text{cur}}$ along the channel dimension, respectively convolving the long-term memory and the current BEV features:
\begin{align}\label{RNN_fusion}
    \overline{H}_{i} &= \Big[f_{\text{sample}}(\overline{H}_{i-1},P_{i,i-1}); B_{i}\Big] * \Big[\mathbf{V}_{\text{mem}}; \mathbf{V}_{\text{cur}}\Big]\\
    &= f_{\text{sample}}(\overline{H}_{i-1},P_{i,i-1}) * \mathbf{V}_{\text{mem}} + B_{i} * \mathbf{V}_{\text{cur}} \notag \\
                     &= f_{\text{sample}}(\overline{H}_{i-2},P_{i,i-2}) * \mathbf{V}_{\text{mem}} * \mathbf{V}_{\text{mem}} + \notag \\ & \quad f_{\text{sample}}(B_{i-1}, P_{i,i-1}) * \mathbf{V}_{\text{cur}} * \mathbf{V}_{\text{mem}} + B_{i} * \mathbf{V}_{\text{cur}} \notag \\
                     &\triangleq f_{\text{sample}}(\overline{H}_{i-2},P_{i,i-2}) * \mathbf{V}_{\text{mem}}^{2} \!+\! f_{\text{sample}}(B_{i-1}, P_{i,i-1}) * \notag \\
                     &\ \quad \mathbf{V}_{\text{cur}} * \mathbf{V}_{\text{mem}} + B_{i} * \mathbf{V}_{\text{cur}} \notag \\
                     &= \sum_{j=1}^{i} f_{\text{sample}}(B_{j},P_{i,j}) * \mathbf{V}_{\text{cur}} * \mathbf{V}_{\text{mem}}^{i-j}.
\end{align}
Here, $B * \mathbf{V}^{n}$ denotes convolution of $B$ with convolution kernel $\mathbf{V}$ repeating $n$ times.

Comparing Eq.~\ref{RNN_fusion} to Eq.~\ref{sliding_window_fusion}, it is seen that the two fusion styles have similar formulations of summing the convolved BEV feature of history frames. This may be the reason both the recurrent paradigm VideoBEV and the parallel paradigm SOLOFusion \cite{SOLOFusion23} can benefit from the long-term temporal information.
However, from the final derivation, we can see that the convolution kernel $\mathbf{V}_{\text{mem}}$ for the $i$-th frame in recurrent style fusion is computed repeatedly $i-j$ times (\emph{i.e.}, the fusion time interval) when fusing with the $j$-th history frame. 
Thus, the recurrent style fusion is aware of the time interval for every history frame.
In contrast, the sliding window fusion kernel $\mathbf{U}_j$ for fusing the $j$-th history frame is computed once for all $j\in\{1,\dots, i\}$.
As a result, it treats every history frame equally without the recurrent syle explicit time interval modeling.
Besides, the sliding window style fusion only fuses the history $k-1$ frames to current frames, while the recurrent style fusion fuses all the history frames, facilitating better 3D perception.

\begin{figure}[t]
    \begin{center}
    \vspace{-0.2in}
    \includegraphics[width=0.6\linewidth]{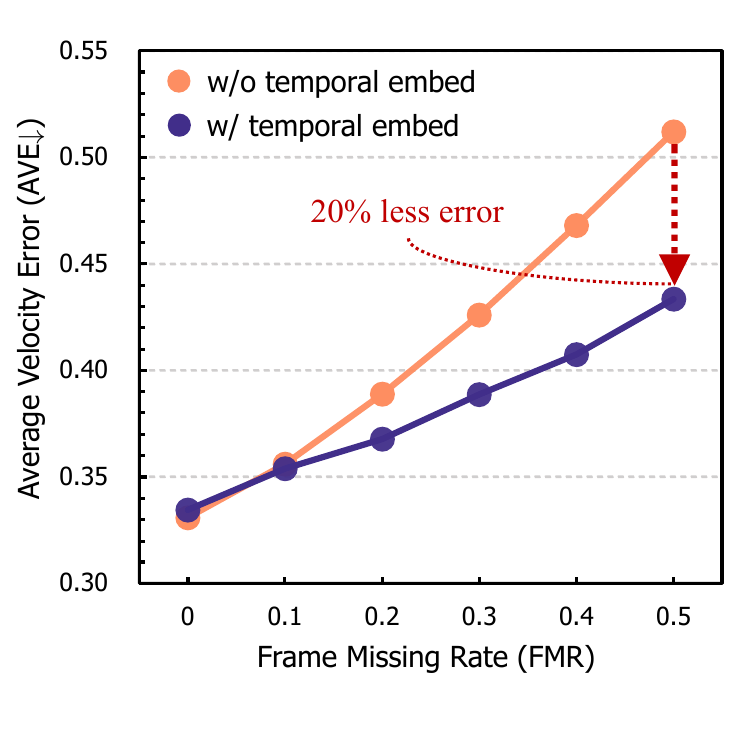}
    \vspace{-0.3in}
    \caption{\textbf{Average velocity error (AVE$\downarrow$) versus frame missing rate (FMR)}. Without the proposed temporal embedding, the AVE is dramatically high when frames are missed, and this issue is substantially mitigated when using the proposed temporal embedding.} \label{fig:frame_lost}
    \end{center}
    \vspace{-0.2in}
\end{figure}

\subsection{Temporal Embedding}
\label{sec:time_embedding}

Generally, the time interval between two adjacent frames in a video sequence is fixed, \emph{e.g.}, 0.5s between two key-frames on nuScenes. However, this can not be guaranteed in the complex real scenes. We empirically find that the missed frames can dreadfully hurt motion estimation. As shown in Fig.~\ref{fig:frame_lost}, the average error of predicted velocity (AVE) becomes dramatically high when the frame missing rate is high. Thus, besides the BEV feature fusion, we propose a temporal embedding module to fuse the time interval between two adjacent frames for stable 3D perception. The temporal embedding module is also designed in a recurrent fashion. Taking $\Delta t_{i} = t_{i} - t_{i-1}$ as the time interval between the $i$-th frame and $i-1$-th frame, the formulation of temporal embedding $\overline{E_{i}}$ for $i$-th frame is as follows:
\begin{align}
    E_{i} &= e(\Delta t_i \cdot \mathbf{1}), \\
    \overline{E}_{i} &= \big[~\overline{E}_{i-1}; E_{i}\big] * K.
\end{align}
Here, $\mathbf{1}$ is the all-one matrix with the same spatial size as $\overline{H}_{i}$, $e(\cdot)$ is the temporal embedding function, which consists of two convolutional layers. $K$ is the convolution kernel for the recurrent fusion. The fused temporal embedding is fed into the velocity head for robust velocity prediction. 

\subsection{Video Inference}
During inference, each frame in the video sequence is evaluated chronologically. The long-term BEV memory feature is initialized with zero. When a new frame comes, the BEV feature is first fused with the memory, based on which the memory is updated and the 3D perceptron is conducted. As a result, the overhead of VideoBEV is consistently low with longer video inputs (see Fig.~\ref{fig:efficiency_compare}). 

\section{EXPERIMENTS}

\subsection{Experimental Setting}

\begin{table*}[t]
    \small
    \begin{center}\renewcommand{\arraystretch}{1.01}
		\caption{\textbf{Comparison results on 3D detection on the nuScenes \texttt{val} set}. All methods in the table are trained with CBGS. \#Frames denotes the number of frames used during training.} \label{table:comparison_to_sota_detection_val_set}
        \vspace{-0.1in}
        \resizebox{0.86\linewidth}{!}{
		\begin{tabular}{l|c|c|c|c|c|c@{\hspace{1.0\tabcolsep}}c@{\hspace{1.0\tabcolsep}}c@{\hspace{1.0\tabcolsep}}c@{\hspace{1.0\tabcolsep}}c}
        \toprule
            \textbf{Method} & \textbf{Backbone} & \textbf{Image Size} & \textbf{\#Frames} & \textbf{mAP}$\uparrow$  &\textbf{NDS}$\uparrow$  & \textbf{mATE}$\downarrow$ & \textbf{mASE}$\downarrow$   &\textbf{mAOE}$\downarrow$   &\textbf{mAVE}$\downarrow$   &\textbf{mAAE}$\downarrow$  \\
            \midrule
            BEVDet~\cite{BEVDet21} & ResNet50 & 256 $\times$ 704 & 1 & 0.298 & 0.379 & 0.725 & 0.279 & 0.589 & 0.860 & 0.245 \\
            PETR~\cite{PETR22} & ResNet50 & 384 $\times$ 1056 & 1 & 0.313 & 0.381 & 0.768 & 0.278 & 0.564 & 0.923 & 0.225 \\ 
            BEVDet4D~\cite{BEVDet4D22} & ResNet50 & 256 $\times$ 704 & 2 & 0.322 & 0.457 & 0.703 & 0.278 & 0.495 & 0.354 & 0.206 \\ 
            BEVDepth~\cite{BEVDepth23} & ResNet50 & 256 $\times$ 704 & 2 & 0.351 & 0.475 & 0.639 & \textbf{0.267} & 0.479 & 0.428 & 0.198 \\
            STS~\cite{STS22} & ResNet50 & 256 $\times$ 704 & 2 & 0.377 & 0.489 & 0.601 & 0.275 & 0.450 & 0.446 & 0.212 \\
            BEVStereo~\cite{BEVStereo23} & ResNet50 & 256 $\times$ 704 & 2 & 0.372 & 0.500 & 0.598 & 0.270 & 0.438 & 0.367 & 0.190 \\ %
            AeDet~\cite{aedet22} & ResNet50 & 256 $\times$ 704 & 2 & 0.387 & 0.501 & 0.598 & 0.276 & 0.461 & 0.392 & 0.196 \\
            SOLOFusion~\cite{SOLOFusion23} & ResNet50 & 256 $\times$ 704 & 17 & \textbf{0.427} & 0.534 & 0.567 & 0.274 & 0.511 & \textbf{0.252} & \textbf{0.188} \\
            \rowcolor[gray]{.95}
            VideoBEV & ResNet50 & 256 $\times$ 704 & 8 & 0.422 & \textbf{0.535} & \textbf{0.564} & 0.276 & \textbf{0.440} & 0.286 & 0.198 \\
            \bottomrule
        \end{tabular}
     }
    \end{center}
    \vspace{-0.2in}
\end{table*}

\begin{table*}
\small
    \begin{center}\renewcommand{\arraystretch}{1.01}
		\caption{\textbf{Comparison results on 3D detection on the nuScenes \texttt{test} set}. TTA denotes test time augmentation strategy. ${}^\dag$ denotes results using future frames during training and inference, and ${}^\ddagger$ denotes results from the official nuScenes leaderboard.} \label{table:comparison_to_sota_detection_test_set}
        \vspace{-0.1in}
        \resizebox{0.86\linewidth}{!}{
        \begin{tabular}{l|c|c|c|c|c|c@{\hspace{1.0\tabcolsep}}c@{\hspace{1.0\tabcolsep}}c@{\hspace{1.0\tabcolsep}}c@{\hspace{1.0\tabcolsep}}c}
        \toprule
        \textbf{Method} & \textbf{Backbone} & \textbf{Image Size} & \textbf{TTA} & \textbf{mAP}$\uparrow$  &\textbf{NDS}$\uparrow$  & \textbf{mATE}$\downarrow$ & \textbf{mASE}$\downarrow$   &\textbf{mAOE}$\downarrow$   &\textbf{mAVE}$\downarrow$   &\textbf{mAAE}$\downarrow$  \\
        \midrule
        FCOS3D~\cite{FCOS3D} & R101-DCN & 900 $\times$ 1600 & \CheckmarkBold & 0.358 & 0.428 & 0.690 & 0.249 & 0.452 & 1.434 & 0.124 \\
        DETR3D~\cite{DETR3D21} & V2-99 & 900 $\times$ 1600 & \CheckmarkBold & 0.412 & 0.479 & 0.641 & 0.255 & 0.394 & 0.845 & 0.133 \\
        UVTR~\cite{UVTR22} & V2-99 & 900 $\times$ 1600 & \XSolidBrush & 0.472 & 0.551 & 0.577 & 0.253 & 0.391 & 0.508 & 0.123 \\
        BEVFormer~\cite{BEVFormer22} & V2-99 & 900 $\times$ 1600 & \XSolidBrush & 0.481 & 0.569 & 0.582 & 0.256 & 0.375 & 0.378 & 0.126 \\
        BEVDet4D~\cite{BEVDet4D22} & Swin-B & 900 $\times$ 1600 & \CheckmarkBold & 0.451 & 0.569 & 0.511 & \textbf{0.241} & 0.386 & 0.301 & 0.121 \\
        PolarFormer~\cite{PolarFormer22} & V2-99 & 900 $\times$ 1600 & \XSolidBrush & 0.493 & 0.572 & 0.556 & 0.256 & 0.364 & 0.439 & 0.127 \\
        PETRv2~\cite{PETRV2} & RevCol  & 640 $\times$ 1600 & \XSolidBrush & 0.512 & 0.592 & 0.547 & 0.242 & 0.360 & 0.367 & 0.126 \\
        HoP-BEVFormer~\cite{zong2023temporal} & V2-99 & 640 $\times$ 1600 & \XSolidBrush & 0.517 & 0.603 & 0.501 & 0.245 & 0.346 & 0.362 & \textbf{0.105} \\
        BEVDepth~\cite{BEVDepth23} & ConvNeXt-B & 640 $\times$ 1600 & \XSolidBrush & 0.520 & 0.609 & 0.445 & 0.243 & 0.352 & 0.347 & 0.127 \\
        BEVStereo~\cite{BEVStereo23} & V2-99 & 640 $\times$ 1600 & \XSolidBrush & 0.525 & 0.610 & \textbf{0.431} & 0.246 & 0.358 & 0.357 & 0.138 \\
        AeDet \cite{aedet22} & ConvNeXt-B & 640 $\times$ 1600 & \CheckmarkBold & 0.531 & 0.620 & 0.439 & 0.247 & \textbf{0.344} & 0.292 & 0.130 \\
        SOLOFusion~\cite{SOLOFusion23} & ConvNeXt-B & 640 $\times$ 1600 & \XSolidBrush & 0.540 & 0.619 & 0.453 & 0.257 & 0.376 & 0.276 & 0.148 \\
        \rowcolor[gray]{.95}
        VideoBEV & ConvNeXt-B & 640 $\times$ 1600 & \XSolidBrush & \textbf{0.554} & \textbf{0.629} & 0.457 & 0.249 & 0.381 & \textbf{0.266} & 0.132 \\
        \midrule
        BEVFormer V2${}^\dag$~\cite{BEVFormerV2} & InternImage-B & 640 $\times$ 1600 & \XSolidBrush & 0.540 & 0.620 & 0.488 & 0.251 & 0.335 & 0.302 & \textbf{0.122} \\
        BEVFormer V2${}^\dag$~\cite{BEVFormerV2} & InternImage-XL & 640 $\times$ 1600 & \XSolidBrush & 0.556 & 0.634 & 0.456 & 0.248 & \textbf{0.317} & 0.293 & 0.123 \\
        BEVFormer V2 Opt${}^{\dagger\ddagger}$ & InternImage-XL & 640 $\times$ 1600 & \XSolidBrush & 0.580 & 0.648 & 0.448 & 0.262 & 0.342 & 0.238 & 0.128 \\
        BEVDet-Gamma${}^{\dagger\ddagger}$ & Swin-B & 640 $\times$ 1600 & \CheckmarkBold & 0.586 & 0.664 & 0.375 & \textbf{0.243} & 0.377 & 0.174 & 0.123 \\
        HoP-BEVDet4D${}^{\dagger}$~\cite{zong2023temporal} & ViT-L & 640 $\times$ 1600 & \XSolidBrush & \textbf{0.624} & textbf{0.685} & \textbf{0.367} & 0.249 & 0.354 & \textbf{0.171} & 0.131 \\
        \rowcolor[gray]{.95}
        VideoBEV${}^\dag$ & ConvNeXt-B & 640 $\times$ 1600 & \XSolidBrush & 0.592 & 0.670 & 0.385 & 0.246 & 0.323 & 0.174 & 0.137 \\
        \bottomrule
    \end{tabular}
    }
    \end{center}
    \vspace{-0.2in}
\end{table*}

\paragraph{Dataset} We use nuScenes dataset~\cite{nuScenes20} for experimental evaluations. It contains 1,000 autonomous driving scenes with around 20 seconds per scene, which is split into 850 scenes for training (\texttt{train}) or validation (\texttt{val}) and 150 for testing (\texttt{test}). 
Six camera images from different perspectives are provided in each frame of the camera data.

\paragraph{Evaluation Metric} We use four commonly used tasks for autonomous driving systems, as stated below. 
We use the typical evaluation criteria for \textbf{3D objection detection} and report the mean Average Precision (mAP) and nuScenes detection score (NDS).
The 3D attributes of translation, scale, orientation, velocity, and attribute are evaluated using the mean Average Translation Error (mATE), mean Average Scale Error (mASE), mean Average Orientation Error (mAOE), mean Average Velocity Error (mAVE), and mean Average Attribute Error (mAAE), respectively. 
The Mean Intersection over Union (mIoU) of the drivable area, the lane, and the vehicle is reported following LSS~\cite{LSS20} for the purpose of \textbf{map segmentation} evaluation.
For object \textbf{tracking} evaluation, we report the average multi-object tracking accuracy (AMOTA), the average multi-object tracking precision (AMOTP), the recall (RECALL), and the multi-object tracking accuracy (MOTA) following the standard assessment metrics. 
For object \textbf{motion prediction} evaluation, we report the minimum Average Displacement Error (minADE), minimum Final Displacement Error (minFDE), Miss Rate (MR), and the End-to-end Prediction Accuracy (EPA) following ViP3D~\cite{ViP3D22}.

\paragraph{Implementation Details}
We conduct our experiments based on the BEVDepth~\cite{BEVDepth23} and BEVStereo~\cite{BEVStereo23}. The learning rate, optimizer, and data augmentation are the same as BEVDepth~\cite{BEVDepth23}. Unless otherwise specified, we use ResNet50~\cite{ResNet16} pre-trained on ImageNet~\cite{ImageNet09} as the image backbone and SECOND FPN~\cite{SECOND18} as the image neck. The size of the BEV feature in all of our experiments is 128 $\times$ 128. The perception ranges are [-51.2m, 51.2m] for the $X$ and $Y$ axis, and the resolution of each BEV grid is 0.8m.

\subsection{Comparison to Prior Arts}
\noindent\textbf{3D Detection}~ To fairly compare with existing SOTAs, we use the ResNet-50, ResNet-101, and ConvNext-Base as backbone respectively. The main results on Nuscenes \texttt{val} and \texttt{test} sets are shown in Tab.~\ref{table:comparison_to_sota_detection_val_set} and Tab.~\ref{table:comparison_to_sota_detection_test_set}. On \texttt{val} set, with the ResNet-50 backbone, the VideoBEV achieves comparable results to SOLOFusion \cite{SOLOFusion23} with fewer frames for training. On the \texttt{test} set, our VideoBEV achieves 55.4\% mAP and 62.9\% NDS without bells and whistles, outperforming all previous methods without the utilization of future frames. Furthermore, our VideoBEV can extend to future frames fusion in the offboard mode, where our method still surpasses most existing methods, including BEVFormer V2 which uses a heavier backbone network (\emph{i.e.}, InternImage-XL~\cite{InternImage23}). These strong results clearly demonstrate the effectiveness of VideoBEV for fusing long-term temporal information. 

\begin{table}[t]
    \small
    \begin{center}\renewcommand{\arraystretch}{1.01}
		\caption{\textbf{Comparison results on map segmentation on the nuScenes \texttt{val} set}. ${}^\dagger$ denotes our baseline method.}
        \vspace{-0.1in}
        \label{table:comparison_to_sota_segmentation}
        \resizebox{0.86\linewidth}{!}{
		\begin{tabular}{l|ccc}
        \toprule
            \textbf{Method} & \textbf{mIoU-Drivable}$\uparrow$ & \textbf{mIoU-Lane}$\uparrow$ & \textbf{mIoU-Vehicle}$\uparrow$  \\
            \midrule
            LSS~\cite{LSS20} & 0.729 & 0.200 & 0.321 \\
            FIERY~\cite{FIERY21} & - & - & 0.382 \\
            M$^{2}$BEV~\cite{M2BEV22} & 0.759 & 0.380 & - \\
            BEVFormer~\cite{BEVFormer22} & 0.775 & 0.239 & 0.467 \\
            UniAD~\cite{UniAD22} & 0.691 & 0.313 & - \\
            BEVDepth$^\dagger$~\cite{BEVDepth23} & 0.816 & 0.453 & 0.460 \\
            \rowcolor[gray]{.95}
            VideoBEV & \textbf{0.827} & \textbf{0.461} & \textbf{0.486} \\
            \bottomrule
        \end{tabular}
     }
    \end{center}
    \vspace{-0.3in}
\end{table}

\begin{table}[t]
    \small
    \begin{center}\renewcommand{\arraystretch}{1.01}
		\caption{\textbf{Comparison results on 3D object tracking on the nuScenes \texttt{test} set}. ${}^\dagger$ denotes our baseline method.}
        \vspace{-0.1in}
        \label{table:comparison_to_sota_tracking}
        \resizebox{0.86\linewidth}{!}{
		\begin{tabular}{l|ccccc}
        \toprule
            \textbf{Method} & \textbf{AMOTA}$\uparrow$ & \textbf{AMOTP}$\downarrow$ & \textbf{RECALL}$\uparrow$ & \textbf{MOTA}$\uparrow$ \\
            \midrule
            CenterTrack~\cite{CenterTrack20} & 0.046 & 1.543 & 23.3\% & 0.043 \\
            DEFT~\cite{DEFT21} & 0.177 & 1.564 & 33.8\% & 0.156 \\
            Time3D~\cite{Time3D22} & 0.210 & 1.360 & - & 0.173 \\
            QD3DT~\cite{QuasiDense23} & 0.217 & 1.550 & 37.5\% & 0.198 \\
            TripletTrack~\cite{TripletTrack22} & 0.268 & 1.504 & 40.0\% & 0.245 \\
            MUTR3D~\cite{MUTR3D22} & 0.270 & 1.494 & 41.1\% & 0.245 \\
            PolarDETR~\cite{PolarDETR22} & 0.273 & 1.185 & 40.4\% & 0.238 \\
            UniAD~\cite{UniAD22} & 0.359 & 1.320 & 46.7\% & - \\
            SRCN3D~\cite{SRCN3D22} & 0.398 & 1.317 & 53.8\% & 0.359 \\
            CC-3DT~\cite{CC3DT22} & 0.410 & 1.274 & 53.8\% & 0.357 \\
            PF-Track~\cite{PFTrack23} & 0.434 & 1.252 & 53.8\% & 0.378 \\
            QTrack$^\dagger$~\cite{QTrack22} & 0.480 & 1.107 & 56.9\% & 0.431 \\
            UVTR~\cite{UVTR22} & 0.519 & 1.125 & 59.9\% & 0.447 \\
            Sparse4D~\cite{Sparse4D22} & 0.519 & 1.078 & \textbf{63.3\%} & 0.459 \\
            \rowcolor[gray]{.95}
            VideoBEV & \textbf{0.548} & \textbf{0.983} & 63.1\% & \textbf{0.475} \\
            \bottomrule
        \end{tabular}
     }
    \end{center}
    \vspace{-0.2in}
\end{table}

\begin{table}[t]
    \small
    \begin{center}\renewcommand{\arraystretch}{1.01}
		\caption{\textbf{Comparison to existing work on prediction on the nuScenes \texttt{val} set}. ${}^\dagger$ denotes our baseline method.}
        \vspace{-0.1in}
        \label{table:comparison_to_sota_prediction}
        \resizebox{0.86\linewidth}{!}{
		\begin{tabular}{l|cccc}
        \toprule
            \textbf{Method} & \textbf{minADE} (m)$\downarrow$ & \textbf{minFDE} (m)$\downarrow$ & \textbf{MR}$\downarrow$ & \textbf{EPA} $\uparrow$ \\
            \midrule
            PnPNet-vision~\cite{PnPNet20} & 2.22 & 3.17 & 0.272 & 0.193 \\
            ViP3D~\cite{ViP3D22} & 2.05 & 2.84 & 0.246 & 0.226 \\
            PIP~\cite{PIP20} & 1.23 & 1.75 & 0.195 & 0.258 \\
            UniAD~\cite{UniAD22} & \textbf{0.71} & 1.02 & 0.151 & 0.456 \\
            BEVDepth$^\dagger$\cite{BEVDepth23} & 1.19 & 1.62 & 0.133 & 0.386 \\
            \rowcolor[gray]{.95}
            VideoBEV & 0.80 & \textbf{0.99} & \textbf{0.067} & \textbf{0.463}\\
            \bottomrule
        \end{tabular}
     }
    \end{center}
    \vspace{-0.2in}
\end{table}

\begin{table}[t]
    \small
    \begin{center}\renewcommand{\arraystretch}{1.01}
		\caption{\textbf{Ablation study on history frames number}. \#Frames denotes used history frames number for training.}
        \vspace{-0.1in}
        \label{table:ablation_number_of_frames}
  \resizebox{0.86\linewidth}{!}{
		\begin{tabular}{c|cc|ccc}
        \toprule
        \textbf{\#Frames} & \textbf{mAP}$\uparrow$  &\textbf{NDS}$\uparrow$ & \textbf{mATE}$\downarrow$ &\textbf{mAOE}$\downarrow$   &\textbf{mAVE}$\downarrow$  \\
        \midrule
        0 & 0.323 & 0.382 & 0.701 & 0.598 & 0.936 \\
        1 & 0.340 & 0.450 & 0.678 & 0.550 & 0.473 \\
        2 & 0.348 & 0.462 & 0.688 & 0.533 & 0.397 \\
        4 & 0.359 & 0.471 & 0.659 & 0.556 & 0.382 \\
        8 & 0.375 & 0.483 & 0.663 & 0.524 & 0.360 \\
        16 & \textbf{0.379} & 0.489 & 0.641 & 0.524 & 0.343 \\
        all & \textbf{0.379} & \textbf{0.492} & \textbf{0.636} & \textbf{0.519} & \textbf{0.331} \\
        \bottomrule
        \end{tabular}
        }
    \end{center}
    \vspace{-0.2in}
\end{table}

\begin{table}[t]
    \small
    \begin{center}\renewcommand{\arraystretch}{1.01}
		\caption{\textbf{Ablation study on combining shallower-layer fusion.} VideoBEV-D and VideoBEV-S represent VideoBEV based on BEVDepth~\cite{BEVDepth23} and BEVStereo~\cite{BEVStereo23}, respectively.}
        \vspace{-0.1in}
        \label{table:combination_with_temporal_stereo}
        \resizebox{0.86\linewidth}{!}{
		\begin{tabular}{l|c@{\hspace{1.0\tabcolsep}}c@{\hspace{1.0\tabcolsep}}|c@{\hspace{1.0\tabcolsep}}c@{\hspace{1.0\tabcolsep}}c@{\hspace{1.0\tabcolsep}}c@{\hspace{1.0\tabcolsep}}c}
        \toprule
        \textbf{Method} & \textbf{mAP}$\uparrow$  &\textbf{NDS}$\uparrow$ & \textbf{mATE}$\downarrow$  &\textbf{mAOE}$\downarrow$   &\textbf{mAVE}$\downarrow$  \\
        \midrule
        BEVDepth~\cite{BEVDepth23} & 0.323 & 0.382 & 0.701 & 0.598 & 0.936 \\ 
        VideoBEV-D & \textbf{0.379} & \textbf{0.492} & \textbf{0.636} & \textbf{0.519} & \textbf{0.331} \\
        \midrule
        BEVStereo~\cite{BEVStereo23} & 0.340 & 0.450 & 0.683 & 0.533 & 0.478 \\ 
        VideoBEV-S & \textbf{0.395} & \textbf{0.502} & \textbf{0.606} & \textbf{0.511} & \textbf{0.344}  \\
        \bottomrule
        \end{tabular}
     }
    \end{center}
    \vspace{-0.2in}
\end{table}

\noindent\textbf{Map Segmentation}~
We evaluate VideoBEV on map segmentation task by simply adding a U-Net-like~\cite{unet15} network for the segmentation of the drivable area, the lane, and the vehicle in BEV. As shown in Tab.~\ref{table:comparison_to_sota_segmentation}, compared to our baseline (single-frame), VideoBEV improves the mIoUs of the three classes by +1.1\%, +0.8\%, and +2.6\%, respectively. VideoBEV surpasses all existing SOTAs, including the BEVFormer \cite{BEVFormer22} and UniAD \cite{UniAD22}. This indicates the temporal information fused by our recurrent fusion module can improve the quality of BEV features for tasks that require dense spatial reasoning.

\noindent\textbf{Object Tracking}~
For 3D multi-object tracking (MOT) task, we employ QTrack~\cite{QTrack22} as our baseline method to generate the trajectories of all predicted 3D objects.
As shown in Tab.~\ref{table:comparison_to_sota_tracking}, VideoBEV achieves the best performance on the nuScenes \texttt{test} set, which outperforms Sparse4D~\cite{Sparse4D22} and UVTR~\cite{UVTR22} by a clear margin of +2.9\% AMOTA. 
Compared to our baseline method QTrack~\cite{QTrack22}, a significant improvement of +6.8\% is observed, demonstrating the superiority and consistent ability to identify objects moving over time.

\noindent\textbf{Object Motion Prediction}~
We also evaluate VideoBEV on the motion prediction task. Inspired by FutureDet~\cite{futuredet}, we first conduct future detection for all target agents in a finite future period (\emph{i.e.}, 6 key-frames in 3s).
Then, we simply utilize the velocity and ti me lag to associate the locations among current and future detection results.
Finally, we take the detection confidence score from the last frame as the score of the corresponding associated motion trajectory.
As shown in Tab.~\ref{table:comparison_to_sota_prediction}, the single-frame BEVDepth baseline using the aforementioned strategy already yields a promising result, outperforming PIP~\cite{PIP20} on all metrics.
Further, by utilizing our efficient temporal fusion strategy of VideoBEV, the SOTA performance is achieved with only a ResNet-50 backbone and 256$\times$704 input resolution.
This demonstrates that our sequential modeling successfully captures the object motion states, which is conducive to future detection for further accurate object motion forecasting.

\begin{figure*}[t]
    \begin{center}
    \vspace{-0.1in}
    \includegraphics[width=0.94\linewidth]{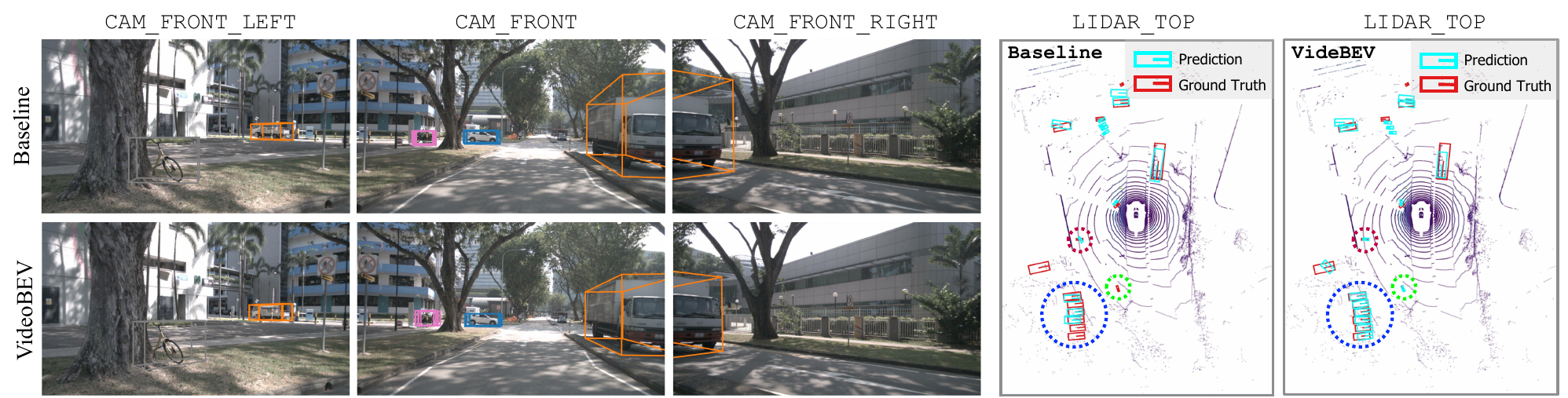}
    \vspace{-0.1in}
    \caption{\textbf{Visualization results of VideoBEV on the nuScenes \texttt{val} set.} We show the predicted 3D box results of single frame baseline and VideoBEV with ResNet-50 backbone in multi-camera images and bird's-eye-view. The results of the baseline involving \textit{false negative}, \textit{incorrect object orientation}, and \textit{inaccurate occluded object identifications} that are fixed by VideoBEV are highlighted with dashed circles in green, purple, and blue, respectively.} \label{fig:vis_results}
    \end{center}
    \vspace{-0.2in}
\end{figure*}

\begin{figure}[t]
    \begin{center}
    \includegraphics[width=0.95\linewidth]{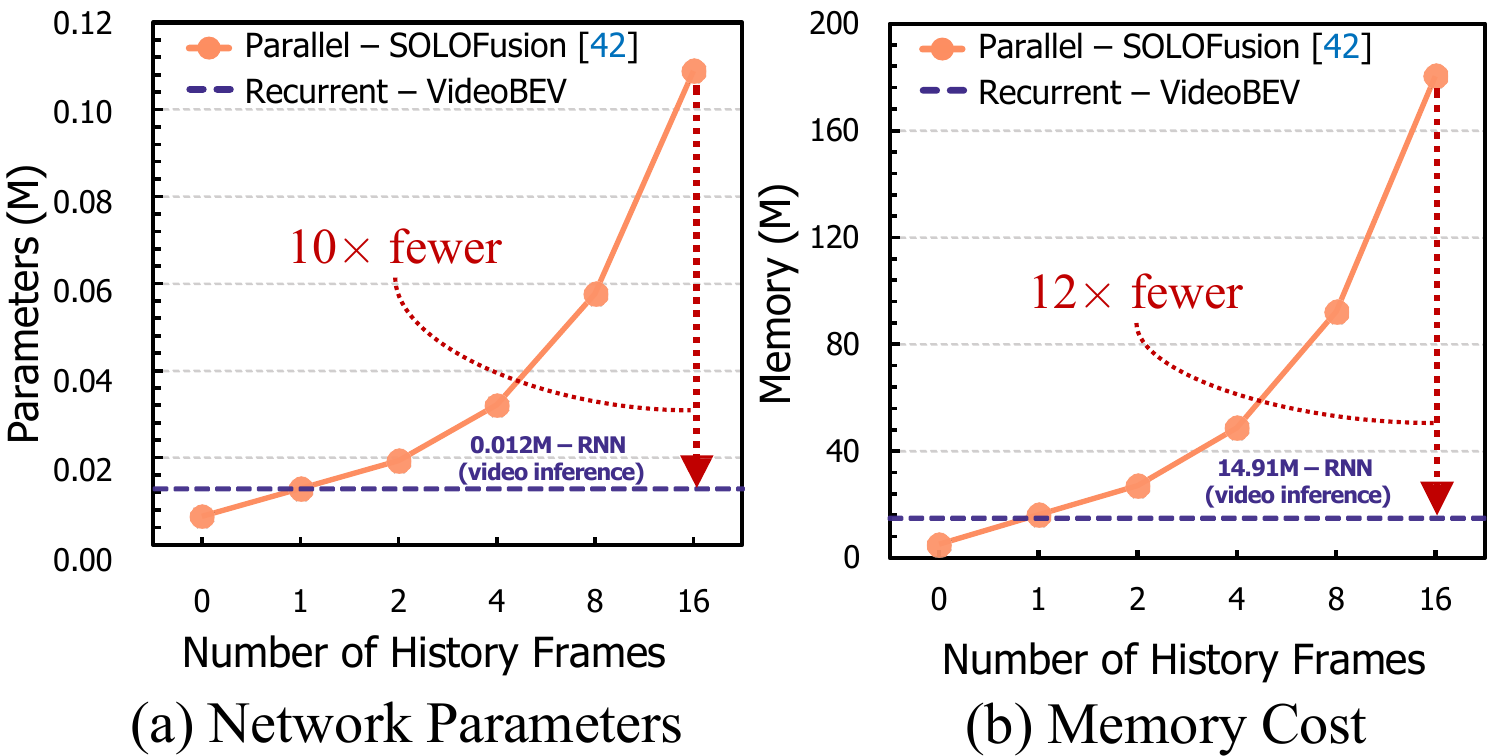}
    \vspace{-0.1in}
    \caption{\textbf{Efficiency comparison of two temporal feature fusion modules}. By comparing with parallel style SOLOFusion~\cite{SOLOFusion23}: (a) Fusion module network parameter; (b) Memory cost of the fusion modules during inference.}\label{fig:efficiency_compare}
    \end{center}
    \vspace{-0.2in}
\end{figure}

\subsection{Ablation Study and Analysis}

\paragraph{Effectiveness of Recurrent Temporal Fusion} To verify the effectiveness of recurrent temporal fusion, we use different numbers of history frames with ResNet-50 backbone for training and testing. As shown in Fig.~\ref{fig:concept_compare}(d) and Tab.~\ref{table:ablation_number_of_frames}, with the increase of used history frames, the mAP and NDS are significantly improved. Specifically, the improvement of VideoBEV with 16 history frames is +5.6\% mAP, +10.7\% NDS over that without temporal fusion. 
When using video inference with all history frames, the performance is further improved compared to the 16 history frames counterpart. This demonstrates that though the frames sampled beyond 16-${th}$ history stamp are far from the current frame, they still contain temporal information useful to the current frame.

\paragraph{Efficiency of Temporal Recurrent Fusion}
VideoBEV recurrently fuses the history BEV feature. Hence, only one BEV feature memory needs to be stored during inference. When a new frame comes, we only need to fuse its BEV feature to the stored one with a lightweight recurrent fusion module. This is efficient for both memory and computation. As shown in Fig.~\ref{fig:efficiency_compare}, when increasing the number of used history frames, the overhead of memory and latency is consistently lower compared to SOLOFusion~\cite{SOLOFusion23}, which is nearly the same as that without any history frame.

\paragraph{Robustness to Missed Frames}
In practice, the frames could sometimes miss, resulting in different time intervals. However, we empirically find that varied time intervals between two adjacent frames can seriously hurt the velocity prediction. We use the frame missing rate (FMR) to study the influence of varied time intervals. As shown in Fig.~\ref{fig:frame_lost}, a higher FMR leads to a higher error in velocity prediction. Specifically, the error of velocity increases by 54.69\% with 50\% FLR compared to that without missed frames. However, when the temporal embedding is used, the error decreases significantly by 25.13\%. This indicates that the temporal embedding correctly encodes the time information for velocity prediction, alleviating the missed frames issue.

\paragraph{Shallower-Layer Fusion}
Short-term temporal fusion is as critical as long-term fusion since it provides more accurate depth estimation through stereo matching~\cite{SOLOFusion23}. 
To study if combining VideoBEV with such short-term fusion could further bring benefits, we implement our VideoBEV based on recently proposed BEVStereo~\cite{BEVStereo23} that leverages short-term fusion.
As shown in Tab.~\ref{table:combination_with_temporal_stereo}, with the shallower-layer temporal fusion (only one history frame is fused), the performance of VideoBEV is significantly improved on mAP and mATE. This implies more advanced temporal fusion on low-level features may further improve the 3D perception, leaving space for future investigation.

\paragraph{Visualization Analysis} 
In Fig.~\ref{fig:vis_results}, we visualize the prediction results of VideoBEV and the single-frame baseline for a qualitative comparison.
By leveraging and fusing long-term temporal information, VideoBEV successfully corrects the wrong predictions caused by commonly met issues like false negatives, wrong orientation estimation, and inaccurate identification for occluded objects. 
It demonstrates the superiority of long-term temporal fusion that may be essential for the understanding of comprehensive driving scenes. 

\section{CONCLUSIONS}
This study investigates a simple recurrent long-term temporal fusion framework based on LSS-based methods for camera-based Bird's-Eye-View 3D perception, dubbed VideoBEV.
Unlike previous works, VideoBEV decouples the recurrent spatiotemporal fusion with a lightweight fusion process. 
Compared to parallel temporal fusion, VideoBEV's resource-efficient recurrent fashion yields a superior computation budget, while enjoying the merits of parallel temporal fusion for long-term information modeling. 
In addition, a dedicated temporal embedding is proposed, which alleviates the frame missing issue in real-world scenarios.
Extensive experiments on diverse BEV 3D perception tasks, including 3D object detection, map segmentation, 3D object tracking, and 3D object motion prediction, are conducted, demonstrating the leading performance of our VideoBEV.
This study reveals that long-term temporal information is essential for comprehensive scene understanding in 3D BEV perception. 
For the first time, we show that a longer-term (\emph{e.g.}, 16 frames in 8s) recurrent temporal fusion brings further benefits for perception accuracy.
This study establishes a new baseline for spatiotemporal BEV 3D perception, and we believe our findings will inspire future research into long-term temporal information fusion for autonomous driving.

\ifCLASSOPTIONcaptionsoff
  \newpage
\fi

{
\bibliographystyle{ieeetr}
\bibliography{sample}
}

\end{document}